\documentclass[runningheads]{llncs}
\usepackage{graphicx}

\usepackage{tikz}
\usepackage{comment}
\usepackage{amsmath,amssymb} % define this before the line numbering.
\usepackage{color}
\usepackage{multirow}
\usepackage{hyperref}
\usepackage[T1]{fontenc}
\usepackage[utf8]{inputenc}
\usepackage[round]{natbib}
\usepackage{afterpage}
\usepackage{subcaption}
\usepackage[width=122mm,left=12mm,paperwidth=146mm,height=193mm,top=12mm,paperheight=217mm]{geometry}

\graphicspath{{./figs/}}

\begin{document}
\pagestyle{headings}
\mainmatter
\def\ECCVSubNumber{100}  % Insert your submission number here

\title{Ramifications of Approximate Posterior Inference for Bayesian Deep Learning in Adversarial and Out-of-Distribution Settings}

% CAMERA READY SUBMISSION
\titlerunning{Ramifications of A.P.I. for BDL in Adversarial \& OoD Settings}
\author{John Mitros\and
Arjun Pakrashi\and
Brian Mac\ Namee}
\authorrunning{J. Mitros et al.}
\institute{School of Computer Science, University College Dublin \\
\email{ioannis.mitros@insight-centre.org}  \\
% \url{http://www.springer.com/gp/computer-science/lncs} 
\email{\{arjun.pakrashi,brian.macnamee\}@ucd.ie}
}
%******************
\maketitle

\begin{abstract}
  Deep neural networks have been successful in diverse discriminative classification tasks, although, they are poorly calibrated often assigning high probability to misclassified predictions. Potential consequences could lead to trustworthiness and accountability of the models when deployed in real applications, where predictions are evaluated based on their confidence scores. Existing solutions suggest the benefits attained by combining deep neural networks and Bayesian inference to quantify uncertainty over the models' predictions for ambiguous data points. In this work we propose to validate and test the efficacy of likelihood based models in the task of out of distribution detection (OoD). Across different datasets and metrics we show that Bayesian deep learning models indeed outperform conventional neural networks but in the event of minimal overlap between in/out distribution classes, even the best models exhibit a reduction in AUC scores in detecting OoD data. We hypothesise that the sensitivity of neural networks to unseen inputs could be a multi-factor phenomenon arising from the different architectural design choices often amplified by the curse of dimensionality. Preliminary investigations indicate the potential inherent role of bias due to choices of initialisation, architecture or activation functions. Furthermore, we perform an analysis on the effect of adversarial noise resistance methods regarding  in and out-of-distribution performance when combined with Bayesian deep learners.
\end{abstract}

\section{Introduction}
Anomaly detection is concerned with the detection of unexpected events. The goal is to effectively detect rare or novel patterns which neither comply to the norm nor follow a specific trend present in the existing data distribution. The task of anomaly detection is often described as open category classification~\citep{liu2018} and anomalies can be referred to as \emph{outliers}, \emph{novelties}, \emph{noise deviations}, \emph{exceptions} or \emph{out-of-distribution} (OoD) examples depending on the context of the application~\citep{shafaei2018}. Anomaly detection raises significant challenges among which some stem from the nature of generated data arising from different sources e.g.~multi-modality from different domains such as manufacturing, industrial, health sectors etc.~\citep{chalapathy2019,hendrycks2018,choi2018}.
\begin{figure}[htpb]
	\centering
	\includegraphics[width=0.92\linewidth]{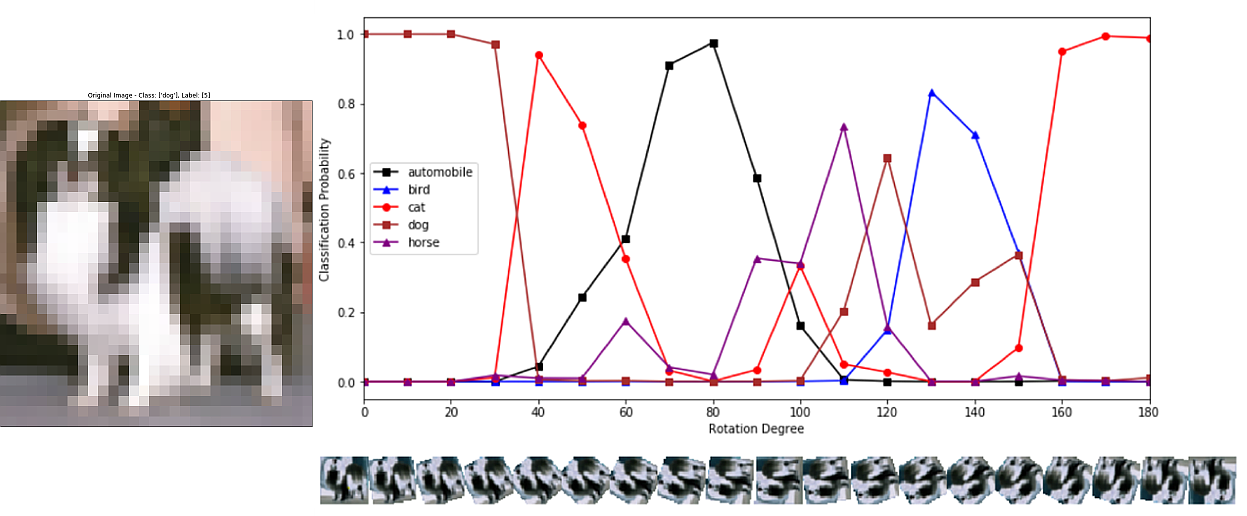}
	\caption{An example of the OoD detection problem. At each angle $0^{o}\le\theta\le180^{o}$ we record the image classifier's response. The classifier is able to predict the true label, \textit{dog}, with approximately 100\% confidence for an angle $\theta\le 30^{o}$ degrees, while it fails to predict the correct label for any rotation angle $\theta> 30^{o}$.
	(Inspired by an image in \citep{gal2016}.)}
~\label{fig:ood_example}
\end{figure}

In this paper we are concerned with a variant of anomaly detection typically referred to as out-of-distribution (OoD) data detection. Figure~\ref{fig:ood_example} shows an illustrative example of an OoD problem. A model is trained to classify images among the following classes: \{\textit{horse}, \textit{dog}, \textit{cat}, \textit{bird}, \textit{automobile}\}. The training procedure involved augmentation, including random rotations. When a new image $\textbf{x}$ is provided containing any of these entities, the model should be able to predict the result and assign a probability to each prediction. In the case of Figure~\ref{fig:ood_example} we record the classifier's response $\hat{y} = f_{\textbf{W}}\left(\textbf{x}\right)$ while rotating an image belonging to the class \emph{dog} in the range of $0^{o}\le\theta\le 180^{o}$ degrees. From Figure~\ref{fig:ood_example} it is evident that the classifier is able to predict the true label $y$, which in this instance is \emph{dog}, with approximately 100\% confidence for angles up to $\theta\le 30^{o}$ degrees, while it fails to predict the correct label for any rotation angle $\theta> 30^{o}$. Moreover, in the cases where the model misclassifies the rotated image, it can assign a very high confidence to the incorrect class membership. For example at $\theta = 40^{o}$ and $\theta = 80^{o}$ the model classifies the rotated image to be a \textit{cat} and an \textit{automobile} respectively with very high confidence.

This uncovers two related issues. First, the example demonstrates the incapability of the model to detect OoD samples previously unseen by it. Second, the classifier misclassifies predictions with high confidence. For instance, the image of the dog was classified as a \textit{cat} or an \textit{automobile} (i.e. black and orange curves in Figure~\ref{fig:ood_example}) with high confidence across different rotation angles. This can potentially lead to problems related with trustworthiness, transparency and privacy of the prediction~\citep{schulam2019}. This phenomenon is not new~\citep{szegedy2013}, and since being identified has captivated significant research effort to understand its underpinnings and provide related solutions.

Many of the proposed solutions have their roots in different fields such as differential privacy, information theory, robust high dimensional statistics, robust control theory and robust optimisation. Recently attention has been concentrated to methods that provide a principled approach to quantifying uncertainty through Bayesian deep learning \citep{gal2016,maddox2019, malinin2018predictive,grathwohl2019your}. In spite of their sophistication, questions remain unanswered regarding the effectiveness of these approaches.

In addition to the OoD prediction problem, there is another issue with which deep neural networks struggle: adversarial noise \citep{akhtar2018threat,ozdag2018adversarial}. For a trained deep neural network model, adversarial noise can be added to an input image in a way such that the model classifies it to a different class from its true one. If this noise is chosen carefully, then the image can appear unchanged to a human observer. As many deep learning based image recognition systems are being deployed in real life, such adversarial attacks can result in serious consequences. Therefore, developing strategies to mitigate against them is an important open problem.

In this paper we describe an empirical evaluation of the effectiveness of Bayesian deep learning for OoD detection on a number of different methods using a selection of image classification datasets. This experiment benchmarks the performance of current state-of-the-art methods, and is aligned in principle to the inquisitive nature of~\cite{goldblum2020_truth_backpropaganda}. Examining why such powerful models exhibit properties of miscalibration and an incapacity to detect OoD samples will allow us to better understand the major implications in critical domain applications since it implies that their predictions cannot be trusted~\citep{hill2019,kumar2019,alemi2018,guo2017,marin2012}. Furthermore, we provide discussions into what we consider to be critical components attributing to this phenomenon which suggest directions for future work, inspired by evidence in recent literature. Finally, we investigate the potential benefit of using adversarial noise defence mechanisms for the OoD detection problem as well as the in-sample adversarial attacks for Bayesian neural networks.

The main contributions of this work include:

\begin{itemize}
  \item A benchmark study demonstrating that Bayesian deep learning methods do indeed outperform point estimate neural networks in OoD detection.
  \item An empirical validation showing that from the considered approaches in this work, \emph{Stochastic Weight Averaging of Gaussian Samples} (SWAG) \citep{maddox2019} is most effective (it achieves the best balance between performance in the OoD detection task and the original image classification task). 
  \item We demonstrate that OoD detection is harder when there exists even minimal overlap between in-distribution and out-of-distribution examples and outlier exposure~\citep{hendrycks2018_outlier_exposure} might not be beneficial in such scenarios.
  \item We investigate the efficacy of adversarial defence techniques and noise robustness showing that randomised smoothing can be quite effective in combination with Bayesian neural networks.
\end{itemize}

This paper is structured as follows: Section \ref{sec:related} describes related work; Section \ref{sec:experiments} describes an experiment to benchmark the ability of Bayesian neural networks to detect OoD examples; Section \ref{sec:adversarial_noise} investigates the effects of adversarial noise defence mechanisms on classification models and the ability of Bayesian neural networks with and without these defences to identify adversarial examples; finally, Section \ref{sec:conclusion} concludes the paper.

\section{Related Work}
\label{sec:related}
The goal of anomaly detection  \citep{chalapathy2019}, is to detect any inputs that do not conform to in-distribution data---referred as \emph{out-of-distribution data} (OoD). While there are approaches that are solely targeted at the detection of anomalies (for example~\citep{chalapathy2019, erfani2016,marchi2015}), classification models that can identify anomalous inputs as part of the classification process are attractive. For instance, it is desirable that a model trained for image classification (such as that described in Figure \ref{fig:ood_example}) would be capable of not only classifying input images accurately, but also identifying that does not belong to any of the classes that it was trained to recognise. Not only can this be useful for detecting ambiguous inputs~\citep{schulam2019}, but it can also be useful for detecting more insidious adversarial examples~\citep{carlini2017}. This type of outlier detection as part of the classification process is most commonly referred to as \emph{OoD detection}.

The majority of OoD detection approaches utilise the uncertainty of classifier's outputs as an indicator to identify whether an input instance is OoD. Such methods are mostly based on density estimation techniques utilising generative models or ensembles. For example, ~\cite{choi2018} presented an ensemble of generative models for density-based OoD detection by estimating epistemic uncertainty of the likelihood, while~\cite{hendrycks2017} focused on detecting malformed inputs, $\textbf{x}$, from their conditional distribution $p(y\vert\boldsymbol{x})$, assigning OoD inputs lower confidence than normal inputs. Hybrid models for uncertainty quantification are also common. For example, the work of~\cite{zhu2017} that describes an end-to-end system using recurrent networks and probabilistic models. The overall architecture is composed of an encoder-decoder recurrent model coupled with a multilayer perceptron (MLP) on top of the encoder allowing to estimate the uncertainty of predictions.

The recent resurgence of interest in combining Bayesian methods with deep neural networks (see \citep{wang2016_bayesian_review}) has given rise to different training mechanisms for classification models that output local likelihood estimates of class membership, providing better estimates of classification uncertainty compared to standard deep neural network (DNN) approaches producing point-estimates. Models whose outputs better represent classification uncertainty can be directly applied to the OoD detection problem---intuitively predictions with high uncertainty suggest OoD examples---and recent approaches that use a Bayesian formulation have claimed to be effective at this. One early example is \emph{MC-Dropout} \citep{gal2016} which views the use of dropout at test time as approximate Bayesian inference. Monte Carlo sampling from the model at test time for different subsets of the weights is used to achieve this. MC-Dropout is based on prior work \citep{damianou2013} which established a relationship between neural networks with dropout and Gaussian Processes (GP). \emph{Stochastic Weight Averaging of Gaussian Samples} (SWAG) \citep{maddox2019}, being an extension of \emph{stochastic weight averaging} (SWA)~\citep{izmailov2018}, is another example of an approximate posterior inference approach that has been shown to be useful for OoD detection. In SWAG, the weights of a neural network are averaged during different SGD iterations~\citep{blei2017} which allows approximates of output distributions to be calculated. 

The \emph{Dirichlet Prior Network} (DPN) \citep{malinin2018predictive} based on recent work focuses explicitly on detecting OoD instances. While alternative approaches using Bayesian principles aim at constructing an implicit conditional distribution with certain desirable properties---e.g., appropriate choice of prior and inference mechanism---DPN strives to explicitly parameterise a distribution using an arithmetic mixture of Dirichlet distributions. DPN also introduces a separate source of uncertainty through an OoD dataset used at training time as a uniform prior over the OoD data (i.e. similar to outlier exposure~\citep{hendrycks2018_outlier_exposure}). Training a DPN model involves optimising the weights of a neural network by minimising the Kullback-Leibler Divergence between in-distribution and out-of-distribution data, where in-distribution data has been modelled with a sharp prior Dirichlet distribution while out-of-distribution data is modelled with a flat prior Dirichlet distribution. 

The \emph{Joint Energy Model} (JEM) \citep{grathwohl2019your} is an alternative approach that uses generative models to improve the calibration of a discriminative model by  reinterpreting an existing neural network architecture as an Energy Based Model (EBM). Specifically they use an EBM \citep{lecun2006tutorial}, which is a type of generative model that parameterises a probability density function using an unnormalised log-density function. It was also shown in \citep{grathwohl2019your} that JEM is effective on OoD problems.

All of these methods perform OoD detection by converting their class membership outputs into a score indicating the likelihood of an out-of-distribution input instance and then applying a threshold to this score. There are four common approaches to calculating these scores:
\begin{itemize}

    \item \emph{Max probability}: This is the maximum value of the softmax outputs. If $\hat{\boldsymbol{y}} = [y_{1},y_{3},\ldots,y_{k}]$ is the prediction, then  $\max_{k} p(y_{k}|\boldsymbol{x})$ is used as the score.
    
    \item \emph{Entropy}: Calculates information entropy $\boldsymbol{H}(Y) = -\sum_{k=1}^{K}p(\hat{y}_{k})\log p(\hat{y}_{k})$ over the class memberships predicted by a model~\citep{malinin2018predictive}.

    \item \emph{Mutual Information}: $\mathbb{E}_{p(\boldsymbol{x},y)}[\frac{p(\boldsymbol{x},y)}{p(\boldsymbol{x})p(y)}]$ measures the amount of information obtained about a random variable X by observing some other random variable Y \citep{tschannen2019mutual}. Here $X$ could describe the entropy of the predictions and $Y$ the entropy of a Dirichlet mixture encompassing the final predictions.

    \item \emph{Differential Entropy}: Also known as the continuous entropy is defined as follows $\boldsymbol{h}(Y) = -\int  p(\hat{y})\log p(\hat{y}) dx$. This is used to measure distributional uncertainty between in-out distributions.
\end{itemize}

These techniques can also be applied to the outputs of a simple deep neural network to perform OoD detection in the same manner.

Despite the resurgence of Bayesian methods for deep learning, there are still a number of challenges that need to be addressed before fully harnessing their benefits for OoD detection.~\cite{foong2019} draws attention to the poorly understood approximations due to computational tractability upon which Bayesian neural networks rely, indicating that common approximations such as factorised Gaussian assumption and MC-Dropout lead to pathological estimates of predictive uncertainty. Posteriors obtained via mean field variation inference (MFVI) \citep{blei2017_review} are unable to represent uncertainty between data clusters but have no issue representing uncertainty outside the data clusters. Similarly, MC-Dropout cannot represent uncertainty in-between data clusters, being more confident in the midpoint of the clusters rather than the centres. In addition,~\cite{wenzel2020} questioned the efficacy of accurate posterior approximation in Bayesian deep learning, demonstrating through MCMC sampling that the predictions induced by a Bayes posterior systematically produced worse results than point estimates obtained from SGD.

Bias in models has been also observed due to the extreme over-parameterised regime of neural networks. Predictions in the vicinity of the training samples are regularised by the continuity of smoothness constraints, while predictions away from the training samples determine the generalisation performance. The different behaviours reflect the inductive bias of the model and training algorithm.~\cite{zhang2020_identity_crisis} showed that shallow networks learn to generalise but as their size increases they are more biased into memorising training samples. Increasing the receptive field of convolutional networks introduces bias towards memorising samples, while increasing the width and number of channels does not. This seems to be in accordance with equivalent findings from~\cite{azulay2019_poor_cnn} identifying that subsampling operations in combination with non-linearities introduce a bias, producing representations that are not shiftable therefore causing the network to lose its invariance properties. Finally,~\cite{fetaya2020_limitations_cond_models} showed that conditional generative models have the tendency to assign higher likelihood to interpolated images due to the class-unrelated entropy indicating that likelihood-based density estimation and robust classification might be at odds with each other.

Prior work has focused on investigating the sensitivity of point estimate DNNs for OoD detection on corrupted inputs as well as comparing local vs global methods (i.e.~ensembles)~\citep{hendrycks2016baseline,chen2020robust,hendrycks2019benchmarking,hendrycks2018deep}. In this work, however, we focus on investigating the effectiveness of local likelihood methods when inputs at test time are considered corrupted with noise by an adversary or simply constitute OoD. In spite of the recent emergence of multiple approaches to the OoD detection problem based on Bayesian methods, and the uncertainty surrounding their effectiveness, no objective benchmark regarding their relative performance at this task currently exists in the literature to best of our knowledge. This paper provides such a benchmark.

To "fool" a deep neural network, examples can be specifically generated in such a way that a specific image is forced to be assigned to a class, from which it does not belong. This is done by adding noise such that it looks identical to the human eye, but it either 1) classifies the example to a specific target class to which the example does not belong (targetted), or 2) to any class other than its original one (untargeted). Such induced noise is called adversarial noise, and  can be generated using several methods \citep{ozdag2018adversarial,akhtar2018threat}. As deep learning based vision systems are being deployed in real world applications, consequently, research in defending against adversarial noise is increasing in popularity and importance. For this pupose, there have been developed several adversarial noise defence mechanisms in the literature \citep{Pang2020Rethinking,xiao2020_kwinners,cohen2019certified}.

\section{Experiment 1: Benchmarking OoD Detection}\label{sec:experiments}

In this experiment, using a number of well-known image classification datasets, we evaluate the ability of four state-of-the-art methods based on Bayesian deep learning---DPN, MC-Dropout, SWAG, and JEM---to detect OoD examples, and compare this with the performance of a standard deep neural network (DNN). 

\subsection{Experimental Setup}
We use a 28 layers wide and 10 layers deep WideResNet~\citep{zagoruyko2016wide} as the DNN model, and this is also the base model used by the other approaches. Whenever it was appropriate and possible for specific datasets, we utilised pre-trained models provided by the original authors to avoid any discrepancies in our results. For instance, \citep{grathwohl2019your} made available a pre-trained JEM model for the CIFAR-10 dataset, and we use this in our experiments rather than training our own. For the remaining models we trained each for 300 epochs using a validation set for hyper-parameter tuning and rolling back to the best network to avoid overfitting.

The optimiser used during the experiments was Stochastic Gradient Descent (SGD) \citep{goodfellow2016} with momentum set to 0.9 and weight decay in the range $[3e^{-4}, 5e^{-4}]$. Additionally, every dataset was split into three distinct sets \{\textit{train, validation, test}\} with augmentations such as random rotation, flip, cropping and pixel distortion applied on the training set. The source code and the pre-trained models are available online.

A significant distinction between DPN and the other approaches used is that DPN uses an idea similar to outlier exposure~\citep{hendrycks2018_outlier_exposure} and it requires two datasets during training: one to represent the in-distribution data and the other to simulate out-of-distribution data. 

Five well-known image classification datasets are used in this experiment:\textit{CIFAR-10} \citep{krizhevsky2009learning}, \textit{CIFAR-100} \citep{krizhevsky2009learning}, \textit{SVHN} \citep{netzer2011reading}, \textit{FashionMNIST} \citep{xiao2017fashion}, and \textit{LSUN}~\citep{yu15lsun} (We use only the \textit{bedroom} scene in this experiment only as OoD data for testing.)

To comprehensively explore the performance of the different models we perform experiments using each dataset to train a model and use all other datasets to measure the ability of the model to perform OoD detection. The ability of each model to perform the image classification task it was trained for, was first evaluated using the test set associated with each dataset. All datasets have balanced class distributions, therefore we use classification accuracy to measure this performance.

To measure the ability of models to recognise OoD examples we make predictions for the test portion of the dataset used to train the model (in-distribution data) and then also make predictions for the test portion of three other datasets (out-of-distribution data): CIFAR-100, SVHN, and LSUN. This means that for each training set we have three different evaluations of OoD detection effectiveness. For example, when SVHN is used as the in-distribution training set, CIFAR-10, LSUN, and CIFAR-100 are used as the out-of-distribution test sets. One of the available out-of-distribution datasets is selected for use at training time for DPN which requires this extra data. Only the training portions  of this dataset is used for this purpose, while the test portions are used for evaluation.

The predictions provided by the models are converted into OoD scores using the four alternative approaches described in Section \ref{sec:related}: \textit{max probability}, \textit{entropy}, \textit{mutual information}, and \textit{differential entropy}. To avoid having to set detection thresholds on these scores we measure the separation between the scores generated for instances on the in-distribution and out-of-distribution test sets using the area under the curve (AUC-ROC) based on the four different approaches for calculating scores. We do this individually for each approach to generate the AUC scores.

\subsection{Results \& Discussion}\label{subsec:exp_1_results}

Table \ref{tab:results} shows the overall performance of each model for the basic in-distribution image classification tasks they were trained for, measured using classification accuracy. These results indicate that, for most cases, the Bayesian methods are performing comparatively to the DNN baseline (the cases where this is  not true will be discussed shortly). 

\begin{table}
\centering
\caption{Accuracy of models on in-distribution dataset image classification tasks.}
\label{tab:results}
\setlength{\tabcolsep}{6pt}
\begin{tabular}{l|llll}
\hline
\multirow{1}{*}{Model} & CIFAR-10 & \multicolumn{1}{c}{SVHN} & \multicolumn{1}{c}{FashionMNIST} & \multicolumn{1}{c}{CIFAR100} \\ \cline{1-5}
DNN & 95.06 & 96.67 & 95.27 & 77.44 \\
DPN & 88.10 & 90.10 & 93.20 & 79.34 \\
MC-Dropout & 96.22 & 96.90 & 95.40 & 78.39 \\
SWAG & 96.53 & 97.06 & 93.80 & 78.61 \\
JEM & 92.83 & 96.13 & 83.21 & 77.86\\ \hline
\end{tabular}
\end{table}

Table~\ref{tab:ood_full_tab} shows the results of the OoD detection experiments. The scores represent AUC-ROC scores calculated using entropy of the model outputs for OoD detection. Values inside parenthesis indicate the percentage improvement with respect to DNN, which is treated as a baseline. The $\uparrow$ indicates an improvement and the $\downarrow$ indicates a degradation. The last row of the table shows the average percentage improvement across the dataset combinations for each approach with respect to the DNN baseline. Similar tables based on OoD scores calculated using max probability, mutual information, and differential entropy are available in the supplementary material. Table~\ref{tab:metric_comparison} summarises the results from these tables and presents the average performance increase with respect to the DNN baseline based on AUC-ROC calculated using each OoD scoring method.

\begin{table}
\begin{centering}
\caption{Out-of-distribution experiment results. Scores are Entropy based AUC-ROC in percentage. The values in bracket are \% improvement of the corresponding algorithm
wrt. DNN, taken as a baseline. An $\uparrow$ indicates improvement and  $\downarrow$ degradation wrt. the baseline (DNN). The asterisks (*) next to each dataset indicates out-distribution datasets used to train DPN. \label{tab:ood_full_tab}}
\par\end{centering}
\centering{}%
\setlength{\tabcolsep}{6pt}
\resizebox{\columnwidth}{!}{ %
\begin{tabular}{ll|lllll}
\hline 
 \multicolumn{2}{c|}{Data}  & (baseline) & \multicolumn{4}{c}{Entropy AUC-ROC score (\% gain wrt. baseline)}\tabularnewline
\cline{3-7} 
In-distribution & OoD & DNN & DPN & MC-Dropout & SWAG & JEM\tabularnewline
\hline 
\multirow{3}{*}{CIFAR-10 } & CIFAR-100* & 86.27 & 85.60 ($\downarrow$0.78\%) & 89.92 ($\uparrow$4.23\%) & 91.89 ($\uparrow$6.51\%) & 87.35 ($\uparrow$1.25\%)\tabularnewline
 & SVHN & 89.72 & 98.90 ($\uparrow$10.23\%) & 96.25 ($\uparrow$7.28\%) & 98.62 ($\uparrow$9.92\%) & 89.22 ($\downarrow$0.56\%)\tabularnewline
 & LSUN & 88.83 & 83.30 ($\downarrow$6.23\%) & 92.04 ($\uparrow$3.61\%) & 95.12 ($\uparrow$7.08\%) & 89.84 ($\uparrow$1.14\%)\tabularnewline
%\hline 
\multicolumn{1}{c}{} & \multicolumn{1}{c|}{} &  &  &  &  & \tabularnewline
%\hline 
\multirow{3}{*}{SVHN} & CIFAR-100 & 93.19 & 99.10 ($\uparrow$6.34\%) & 94.33 ($\uparrow$1.22\%) & 95.97 ($\uparrow$2.98\%) & 92.34 ($\downarrow$0.91\%)\tabularnewline
 & CIFAR-10* & 94.58 & 99.60 ($\uparrow$5.31\%) & 94.97 ($\uparrow$0.41\%) & 96.03 ($\uparrow$1.53\%) & 92.85 ($\downarrow$1.83\%)\tabularnewline
 & LSUN & 92.97 & 99.70 ($\uparrow$7.24\%) & 93.31 ($\uparrow$0.37\%) & 95.71 ($\uparrow$2.95\%) & 91.82 ($\downarrow$1.24\%)\tabularnewline
%\hline 
\multicolumn{1}{c}{} & \multicolumn{1}{c|}{} &  &  &  &  & \tabularnewline
%\hline 
\multirow{3}{*}{FashionMNIST} & CIFAR-100 & 91.20 & 99.50 ($\uparrow$9.10\%) & 93.75 ($\uparrow$2.80\%) & 96.19 ($\uparrow$5.47\%) & 62.79 ($\downarrow$31.15\%)\tabularnewline
 & CIFAR-10* & 94.59 & 99.60 ($\uparrow$5.30\%) & 96.06 ($\uparrow$1.55\%) & 94.28 ($\downarrow$0.33\%) & 64.76 ($\downarrow$31.54\%)\tabularnewline
 & LSUN & 93.34 & 99.80 ($\uparrow$6.92\%) & 97.40 ($\uparrow$4.35\%) & 99.05 ($\uparrow$6.12\%) & 65.38 ($\downarrow$29.96\%)\tabularnewline
%\hline 
\multicolumn{1}{c}{} & \multicolumn{1}{c|}{} &  &  &  &  & \tabularnewline
%\hline 
\multirow{3}{*}{CIFAR-100} & CIFAR-10 & 78.25 & 85.15 ($\uparrow$8.82\%) & 80.70 ($\uparrow$3.13\%) & 84.92 ($\uparrow$8.52\%) & 77.64 ($\downarrow$0.78\%)\tabularnewline
 & SVHN* & 81.52 & 92.64 ($\uparrow$13.64\%) & 85.59 ($\uparrow$4.99\%) & 94.16 ($\uparrow$15.51\%) & 81.22 ($\downarrow$0.37\%)\tabularnewline
 & LSUN & 77.22 & 86.38 ($\uparrow$11.86\%) & 76.58 ($\downarrow$0.83\%) & 87.22 ($\uparrow$12.95\%) & 77.54 ($\uparrow$0.41\%)\tabularnewline
%\hline 
\hline 
\multicolumn{2}{l|}{Avg \% improvement} & & \multicolumn{1}{r}{($\uparrow$6.48\%)} & \multicolumn{1}{r}{($\uparrow$2.76\%)} & \multicolumn{1}{r}{($\uparrow$6.60\%)} & \multicolumn{1}{r}{($\downarrow$7.96\%)}\tabularnewline
\hline 
\end{tabular}}
\end{table}

\clearpage

\begin{figure}[ht!]
	\centering
	\includegraphics[width=0.92\linewidth]{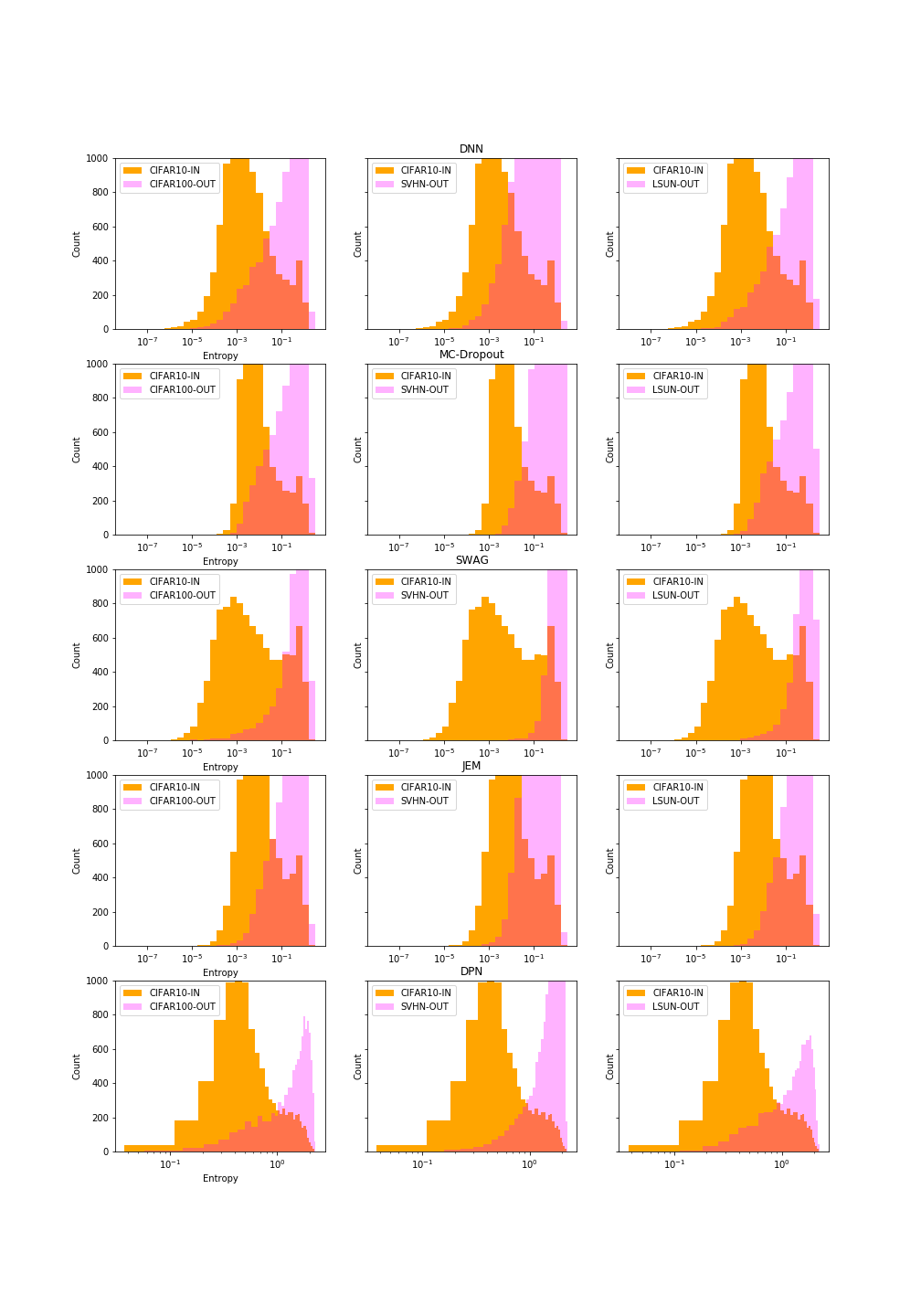}
	\caption{Histograms of in-out distribution across models and datasets. The x-axis represents entropy in log scale.}
~\label{fig:histograms}
\end{figure}

\afterpage{
\begin{table}
\caption{Percent performance increase wrt. baseline (DPN) for all the evaluation
score. The values in parenthesis are the relative ranks of improvement
for the corresponding evaluation score. Last row indicates the improvement
ranking of the algorithms, averaged through the evaluation scores.}\label{tab:metric_comparison}

\begin{centering}
\setlength{\tabcolsep}{6pt}
\begin{tabular}{l|llll}
\hline 
 & DPN & MC-Dropout & SWAG & JEM \\
\hline 
Max prob.     & 5.895\% (2) & 2.618\% (3) & 6.067\% (1) & -8.350\% (4)\tabularnewline
Mutual Info.  & 7.262\% (1) & 6.243\% (2) & 5.337\% (3) & -8.034\% (4)\tabularnewline
Entropy       & 6.480\% (2) & 6.480\% (3) & 6.601\% (1) & -7.960\% (4)\tabularnewline
Diff. Entropy & 9.232\% (1) & 4.386\% (3) & 5.495\% (2) & -10.217\% (4)\tabularnewline
\hline 
Avg. rank & \multicolumn{1}{r}{(1.50)} & \multicolumn{1}{r}{(2.75)} & \multicolumn{1}{r}{(1.75)} & \multicolumn{1}{r}{(4.00)}\tabularnewline
\hline 
\end{tabular}
\par\end{centering}
\end{table}}

It is evident from Table \ref{tab:ood_full_tab} that most of the Bayesian methods--- DPN, MC-Dropout and SWAG---almost always improve OoD detection performance over the DNN baseline. Interestingly, in our experiment JEM consistently performed poorly with respect to the DNN baseline. We attribute this mostly to a failure of robust selection of hyper-parameter combinations. Despite our best efforts we were not able to achieve classification accuracy (above 83\%)  on the FashionMNIST dataset using this approach, and the OoD performance suffered from this. However, even for the datasets in which the JEM models performed well for the image classification task (SVHN and CIFAR-100) its use did not lead to an increase in OoD detection performance \footnote{Even after suggestions kindly provided by~\citep{grathwohl2019your} we found that the model diverged instead of closely approximating its original results. There was another instance we noticed a degradation in performance compared to the original results, it was in the case of DPN when trained with CIFAR-10 for in-distribution and CIFAR-100 for OoD~\citep{malinin2018predictive}, due to the overlap between some classes in CIFAR-10 \& CIFAR-100.}. Overall we found that both JEM and DPN can be quite unstable and extremely sensitive to hyper-parameter choice.

The results in Table \ref{tab:metric_comparison} indicate that among the different evaluation metrics, DPN performs best for OoD detection, with SWAG following closely behind. MC-Dropout is third, but comparatively worse than the other two. All three Bayesian methods on average increase OoD detection performance with respect to the DNN baseline regardless of the approach used to calculate OoD scores. DPN used in combination with differential entropy is especially effective. For other methods the best performance is achieved using entropy as the score for the OoD detection task. This indicates that overall, Bayesian methods improve the performance on OoD detection, with DPN and SWAG achieving similar performance (although JEM does not achieve improvements over the DNN baseline). 

However, it is worth noting from Table \ref{tab:results} that the DPN models do not perform well on the base image classification tasks compared to the other approaches. Therefore, we conclude that of the approaches compared in this experiment SWAG is the most effective as it achieves strong OoD detection performance, without compromising  basic in-distribution classification performance. It also does not require the use of an OoD dataset at training time. Although, if the objective is achieving the best possible OoD detection performance then DPN should be considered instead.

It is also worth mentioning that the performance of models used in combination with the four approaches to calculating OoD scores (described in  Section \ref{sec:related}), varies slightly. The \textit{max probability}, \textit{entropy} and \textit{mutual information} approaches are sensitive to class overlap between in-distribution and out-of-distribution data, while \textit{differential entropy} seems to be more sensitive to the target precision parameter of the Dirichlet distribution, controlling how it is concentrated over the simplex of possible outputs.

\section{Experiment 2: Robustness Analysis}
\label{sec:adversarial_noise}

Having identified in the previous experiment that Bayesian neural network models can be utilised for OoD detection, we proceed by verifying whether they can withstand untargeted adversarial attacks. We design a new experiment including the usual DNN acting as a control baseline and two Bayesian models (\textit{MC-Dropout} and \textit{SWAG}), in order to answer the following questions: 
\begin{enumerate}
    \item Are Bayesian neural networks capable of detecting adversarial examples? 
    \item Could simple defence mechanisms \{\textit{Top-k, Randomised Smoothing, MMLDA}\}, outperform BNN or improve their overall performance? 
\end{enumerate}

\subsection{Experimental Setup}

First, we evaluated the ability of each model against untargeted adversarial examples generated with \emph{Projected Gradient Descent (PGD)} \citep{madry2017towards, ozdag2018adversarial} with the following hyper-parameter values $\epsilon = 0.1$ and $\alpha = 0.01$ for ten iterations.  Second, we introduced three recently proposed defence techniques against adversarial examples and evaluated their efficacy on the (i)\textit{clean test set}, (ii)\textit{adversarially corrupted test set}, and finally on the (iii)\textit{OoD detection task}. The defence methods evaluated against adversarial examples are: \textit{Randomised Smoothing}
~\citep{cohen2019certified}, \textit{Sparsify k-winners take all}~(\emph{Top-k}) \citep{xiao2020_kwinners}, and \textit{Max-Mahalanobis Linear Discriminant Analysis}~(MMLDA) \citep{Pang2020Rethinking}. We use only the CIFAR-10 dataset in these experiments.

In order to train models with MMLDA and RandSmooth we utilised the same number of hyper-parameters and values as they are depicted in the original papers. For instance, for MMLDA we utilised a variance of value 10 for CIFAR-10 in order to compute the Max-Mahalanobis centres on $\mathbf{z}$, where $\mathbf{z}$ denotes the features regarding the penultimate layer of each model $\mathbf{z} = f(x)$. We should also notice the possibility for randomised smoothing to achieve better results at the expense of increased prediction time. A key combination for obtaining successful results is the number of samples $n=55$, or randomised copies for each $x$ in the test set, together with the confidence level $\alpha = 0.001$ (i.e.~there is a 0.001 probability that the answer will be wrong) and standard deviation  $\sigma = 0.56$ of the isotropic Gaussian noise.

\subsection{Results \& Discussion}\label{subsec:exp_2_results}

We first illustrate the effect of adversarial noise on the ability of different model types to perform the underlying CIFAR-10 classification task in the presence of adversarial noise. Table~\ref{tab:ablation_in_dist}
%and Figure \ref{fig:ablation_barplot} 
shows the performance of each model type on the basic CIFAR-10 classification problem on a clean dataset with no adversarial noise. We see that, apart from randomise smoothing, the addition of noise defence strategies does not negatively affect performance at this task. Randomised smoothing had a negative impact on the in-distribution performance across all methods indicating that robustness might be at odds with accuracy~\citep{tsipras2018robustness}. MMLDA and Top-k seem to have a positive effect on DNN and MC-Dropout but not on SWAG.

%\afterpage{
\begin{table}[htp]
\caption{Accuracy on CIFAR-10 clean test set for each defence technique. Values in parenthesis indicate the percentage of abstained predictions.\label{tab:ablation_in_dist}}
\centering
\setlength{\tabcolsep}{6pt}
\begin{tabular}{l|llll}
\hline
Model       & No defence & Top-k & RandSmooth & MMLDA \tabularnewline
\hline
DNN         & 95.06 & 94.52 & 86.25 (13.00) & 95.18 \tabularnewline
MC-Dropout  & 96.22 & 94.43 & 86.98 (13.39) & 95.21 \tabularnewline
SWAG        & 96.53 & 91.73 & 79.68 (20.32) & 91.30\tabularnewline
\hline
\end{tabular}
\end{table}
%}

\begin{figure}[htb]
	\centering
	\includegraphics[width=0.92\linewidth]{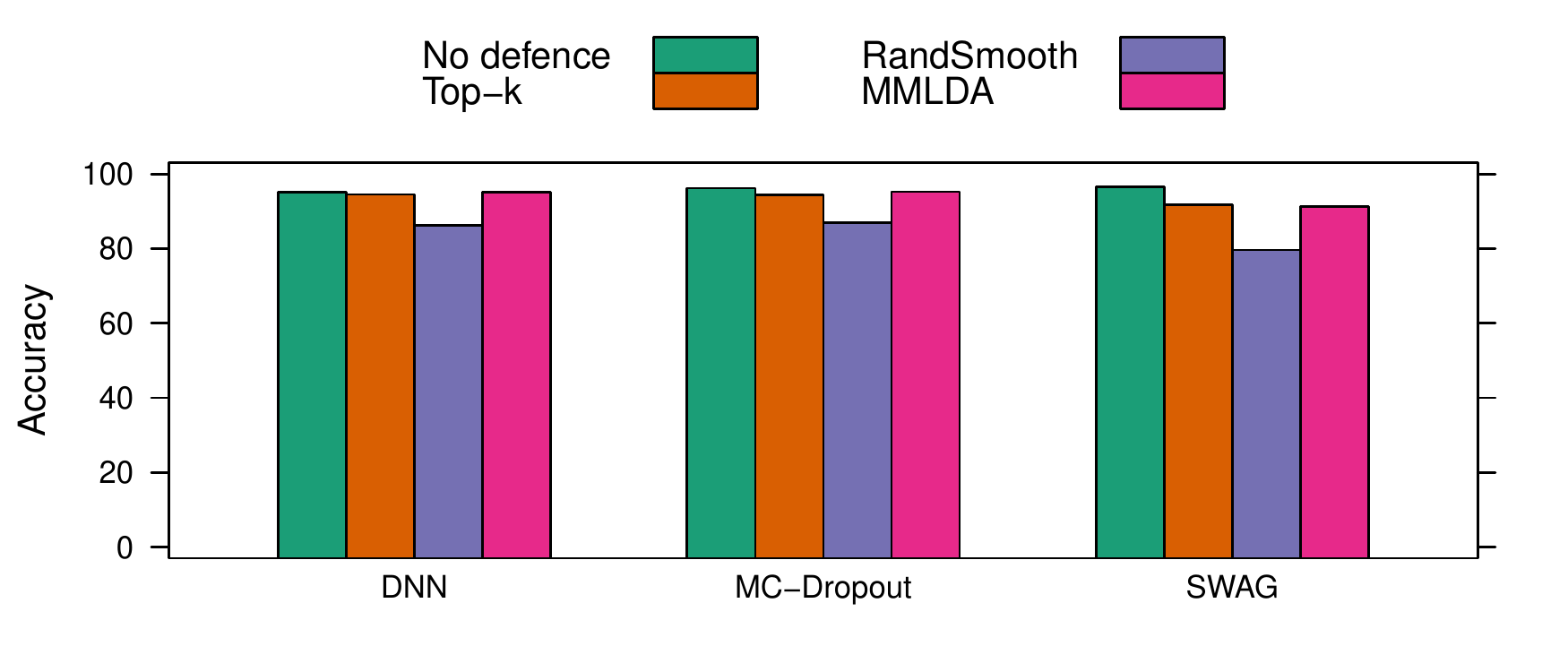}
	\caption{Evaluation of defence techniques vs non-defence on clean test set of CIFAR-10.}
~\label{fig:indist_ablation_comp}
\end{figure}

Table~\ref{tab:adv_noise_exp}
summarises the performance of the same models when adversarial noise is added to the test set. The first column in Table~\ref{tab:adv_noise_exp} corresponds to the accuracy on the adversarially corrupted test set for each model without applying any adversarial defence mechanism. Lower values indicate that the adversarial attack was successful and managed to force the model to misclassify instances with high confidence. When Randomised smoothing is used, the values in parentheses denote the percentage of abstained predictions on the total number of instances from the test set. The impact of adversarial noise is evident in these results. When there is no defence mechanism used then classification accuracy plummets to less than $2\%$ for all models. The addition of defence techniques, especially MMLDA and randomised smoothing indicate improvement in performance.  
\begin{table}[ht]
\caption{Accuracy on CIFAR-10 test set corrupted with adversarial noise.~\label{tab:adv_noise_exp}}
\centering
\setlength{\tabcolsep}{6pt}
\begin{tabular}{l|llll}
\hline
Model & No Defence & Top-k  & RandSmooth & MMLDA \tabularnewline
\hline
DNN & 1.15 & 11.41 & 62.65 (27.75) & 45.71 \tabularnewline
MC-Dropout & 1.94 & 7.28 & 88.85 (17.98) & 47.90 \tabularnewline
SWAG       & 0.55 & 0.79 & 36.80 (20.97) & 47.95 \tabularnewline
\hline
\end{tabular}
\end{table}

\begin{figure}[htb]
	\centering
	\includegraphics[width=0.92\linewidth]{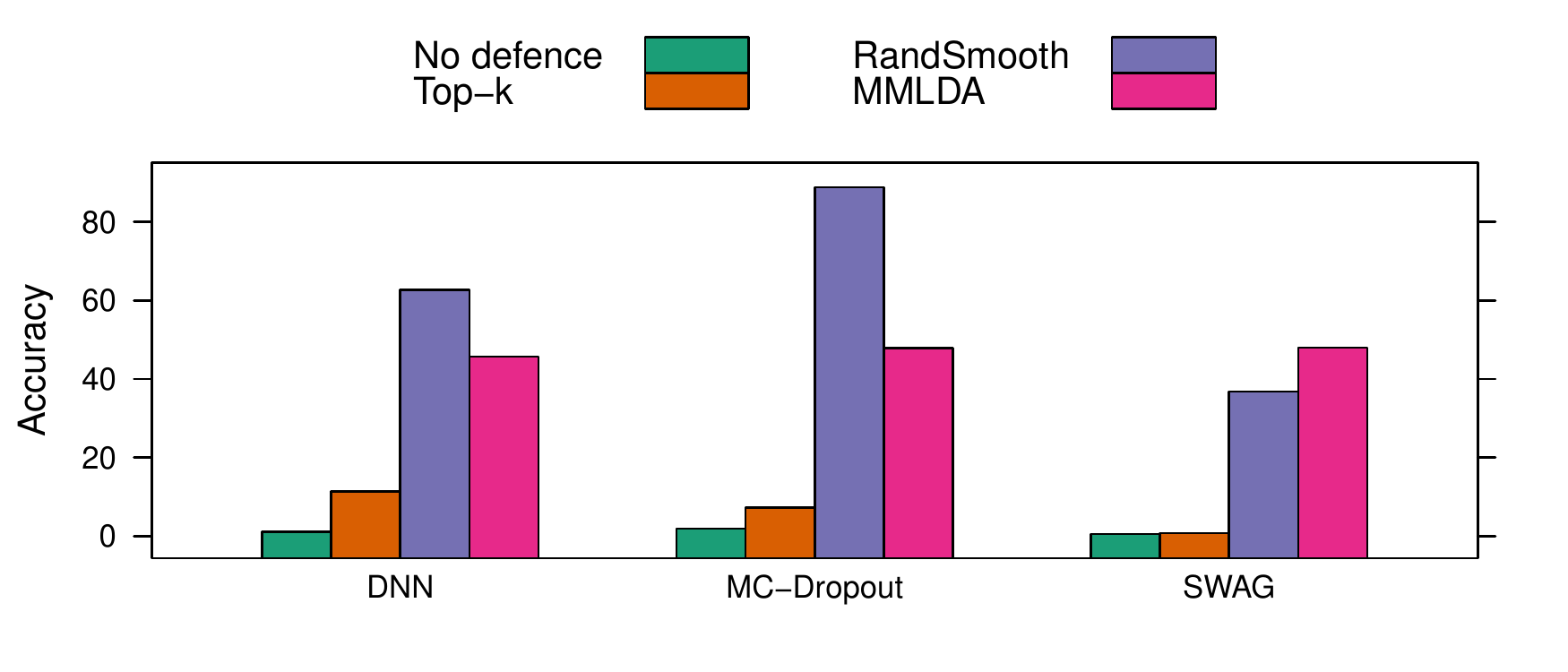}
	\caption{Evaluation of defence techniques vs non-defence on the adversarially corrupted test set of CIFAR-10.}
~\label{fig:noise_comp}
\end{figure}

In addition, Table \ref{tab:adversarial_ood}, shows that model predictions can be used to identify adversarial examples in the same manner that OoD examples were identified by measuring the entropy over the final predictions. It compares the entropy scores generated for an in-distribution test set versus those generated for adversarial examples and uses AUC-ROC scores to measure the ability of a model to distinguish them. First, notice that in line with the results in the previous section the Bayesian models perform better at this task than the baseline DNN when no defence against adversarial examples is used. SWAG, in particular, seems quite effective. When adversarial defence mechanisms are used, the performance increases dramatically. Top-k in particular seems effective in this occasion. \clearpage
\begin{table}[htb]
\caption{Out-of-distribution detection results for all defence methods on clean vs adversarially corrupted CIFAR-10. Scores represent entropy based AUC-ROC in percentage.\label{tab:adversarial_ood}}
\centering
% \resizebox{\columnwidth}{!}{ %
\setlength{\tabcolsep}{6pt}
\begin{tabular}{ll|lll}
\hline 
 \multicolumn{2}{c|}{Data}  & \multicolumn{3}{c}{Entropy AUC-ROC score}\tabularnewline
 \hline
%\cline{3-7} 
In-distribution & OoD & DNN & MC-Dropout & SWAG \tabularnewline
\hline 
CIFAR-10/No defence & CIFAR-10 Adv. & 13.67 & 36.54 & 73.81\tabularnewline
% & & & & \\
CIFAR-10/Topk & CIFAR-10 Adv. & 99.83 & 99.83 & 98.90\tabularnewline
% & & & & \\
CIFAR-10/MMLDA
& CIFAR-10 Adv. & 53.15 & 85.65 & 70.38 \tabularnewline
% & & & & \\
CIFAR-10/RandSmooth
& CIFAR-10 Adv. & 62.68 & 57.06 & 48.29 \tabularnewline
\hline 
\end{tabular} %}
\end{table}

Finally, Table~\ref{tab:ablation_ood} shows whether the defence techniques provide any improvements in regard to the OoD detection task in which the defence techniques are incorporated within the models. The values represent AUC-ROC scores computed based on entropy of the predictions obtained from each model after having introduced the different defence techniques to each model. Again the top-k and MMLDA approaches improve performance over the no defence option.
 %\afterpage{
\begin{table}[ht]
\caption{Out-of-distribution detection results. Scores represent entropy based AUC-ROC in percentage.\label{tab:ablation_ood}}
\centering
% \resizebox{\columnwidth}{!}{ %
\setlength{\tabcolsep}{6pt}
\begin{tabular}{ll|lll}
\hline 
 \multicolumn{2}{c|}{Data}  & \multicolumn{3}{c}{Entropy AUC-ROC score}\tabularnewline
 \hline
%\cline{3-7} 
In-distribution & OoD & DNN & MC-Dropout & SWAG \tabularnewline
\hline 
\multirow{3}{*}{CIFAR-10/Topk}
& CIFAR-100 & 90.59 & 90.45 & 84.72\tabularnewline
& SVHN & 91.20 & 92.62  & 94.61  \tabularnewline
& LSUN & 92.42 & 92.39 & 89.81 \tabularnewline 
& & & & \\
\multirow{3}{*}{CIFAR-10/MMLDA}
& CIFAR-100 & 99.87 & 99.24 & 79.78 \tabularnewline
& SVHN & 99.74 & 99.75 & 84.56 \tabularnewline
& LSUN & 99.93 & 99.67 & 81.72 \tabularnewline 
& & & & \\
\multirow{3}{*}{CIFAR-10/RandSmooth}
& CIFAR-100 & 62.91 & 69.33 & 60.21 \tabularnewline
& SVHN & 42.62 & 63.22 & 66.48 \tabularnewline
& LSUN & 61.94 & 69.10 & 61.23 \tabularnewline
\hline 
\end{tabular} %}
\end{table}
 %}

\section{Conclusion}
\label{sec:conclusion}
This work investigates if Bayesian methods improve OoD detection for deep neural networks used in image classification. To do so, we investigated four recent methods and compared them with a baseline point estimate neural network on four datasets. 

Our findings show that the Bayesian methods overall improve performance at the OoD detection task over the DNN baseline. Of the considered methods examined DPN performed best, with SWAG following closely behind. However, DPN requires additional data during training, is sensitive to hyper-parameter tuning, and led to worse in-distribution performance. Therefore, we conclude that SWAG seems more effective of the approaches examined considering both in and out-distribution performance.

Furthermore we showed that, despite being better than simple DNN models on OoD, Bayesian neural networks do not possess the ability to cope with adversarial examples.  Although adversarial defence techniques overall degrade accuracy on the clean test set, at the same time, they robustify against adversarial examples across models and occasionally improve OoD detection. Randomised smoothing performed the best, followed by MMLDA. In the case of the OoD detection task it seems that Top-k slightly improved performance for DNN while MMLDA improved performance for DNN and MC-Dropout.

The broader scope of our work is not only to identify whether Bayesian methods could be used for OoD detection and to mitigate against adversarial examples, but also to understand how much of the proposed methods are affected by the inductive bias in the choices of model architecture and objective functions. For our future work we look forward to examine these components and their inherent role regarding the sensitivity of models against outliers.

\section*{Acknowledgements}
This research was supported by Science Foundation Ireland (SFI) under Grant number SFI/12/RC/2289\_P2.

\bibliographystyle{unsrtnat}
\bibliography{egbib}
\end{document}